\documentclass[conference]{IEEEtran}
\IEEEoverridecommandlockouts
\usepackage{booktabs}
\usepackage{cite}
\usepackage{amsmath,amssymb,amsfonts}
\usepackage{algorithmic}
\usepackage{comment}
\usepackage{graphicx}
\usepackage[hang,flushmargin]{footmisc}
\usepackage{placeins}
\usepackage{subcaption}
\usepackage{textcomp}
\usepackage[table,xcdraw]{xcolor}
\usepackage{xcolor}
\usepackage{booktabs}
\usepackage{multirow}
\usepackage{siunitx}
\usepackage{titlesec}
\usepackage{hyperref}
\hypersetup{
  colorlinks=true,
  allcolors=blue
}
\def\BibTeX{{\rm B\kern-.05em{\sc i\kern-.025em b}\kern-.08em
    T\kern-.1667em\lower.7ex\hbox{E}\kern-.125emX}}
\usepackage{ifthen}

\newcommand{\ccl}[1]{%
    \ifdim #1 pt < 0pt {\color{red} $#1$} \else $#1$ \fi
}

\usepackage{pifont}

\newcommand{\terr}[1]{{\scriptsize{\ensuremath{{\color{gray}\,\pm{}\,#1}}}}}
\newcommand{\hide}[1]{}
\def\Snospace~{\S{}}

\begin{document}

\title{Hierarchical Multi-Label Classification with Missing Information for Benthic Habitat Imagery}

\author{
\IEEEauthorblockN{
Isaac Xu\IEEEauthorrefmark{1},
Benjamin Misiuk\IEEEauthorrefmark{2,3},
Scott C. Lowe\IEEEauthorrefmark{1,4},
Martin Gillis\IEEEauthorrefmark{1},
Craig J. Brown\IEEEauthorrefmark{5}, and
Thomas Trappenberg\IEEEauthorrefmark{1,*}
}
\vspace{0.2cm}
\IEEEauthorblockA{\IEEEauthorrefmark{1}Faculty of Computer Science, Dalhousie University, Halifax, Nova Scotia, Canada}
\IEEEauthorblockA{\IEEEauthorrefmark{2}Department of Geography, Memorial University of Newfoundland, St. John's, Newfoundland, Canada}
\IEEEauthorblockA{\IEEEauthorrefmark{3}Department of Earth Sciences, Memorial University of Newfoundland, St. John's, Newfoundland, Canada}
\IEEEauthorblockA{\IEEEauthorrefmark{4}Vector Institute, Toronto, Ontario, Canada}
\IEEEauthorblockA{\IEEEauthorrefmark{5}Department of Oceanography, Dalhousie University, Halifax, Nova Scotia, Canada}
}


\maketitle

\renewcommand{\thefootnote}{\fnsymbol{footnote}}
\footnotetext[1]{Corresponding author: \href{mailto:tt@cs.dal.ca}{tt@cs.dal.ca}}
\renewcommand{\thefootnote}{\arabic{footnote}}

\begin{abstract}
In this work, we apply state-of-the-art self-supervised learning techniques on a large dataset of seafloor imagery, \textit{BenthicNet}, and study their performance for a complex hierarchical multi-label (HML) classification downstream task. In particular, we demonstrate the capacity to conduct HML training in scenarios where there exist multiple levels of missing annotation information, an important scenario for handling heterogeneous real-world data collected by multiple research groups with differing data collection protocols.
We find that, when using smaller one-hot image label datasets typical of local or regional scale benthic science projects, models pre-trained with self-supervision on a larger collection of in-domain benthic data outperform models pre-trained on ImageNet. In the HML setting, we find the model can attain a deeper and more precise classification if it is pre-trained with self-supervision on in-domain data.
We hope this work can establish a benchmark for future models in the field of automated underwater image annotation tasks and can guide work in other domains with hierarchical annotations of mixed resolution.\footnote{Code available at: \url{https://github.com/DalhousieAI/benthicnet\_probes} and \url{https://github.com/DalhousieAI/ssl-bentho}}
\end{abstract}


\section{Introduction}

Monitoring and categorizing the changes in seafloor habitats (benthic habitats) is a task of increasing importance due to the pressing need to understand the impact of anthropic disruption via activities such as fishing and climate change. 
An essential datastream for the monitoring of ocean sustainability is photographic imagery of the seafloor. 
In recent years, manually controlled and automated underwater vehicles have permitted a substantial increase the volume of seafloor imagery being collected. 
However, this growing scale of data collection in monitoring seafloor environments renders manual imagery annotation impractical. 
For this reason, automating the processing and classification of benthic photographs is essential.

\begin{figure}[ht]
    \centering
    \includegraphics[width=\columnwidth]{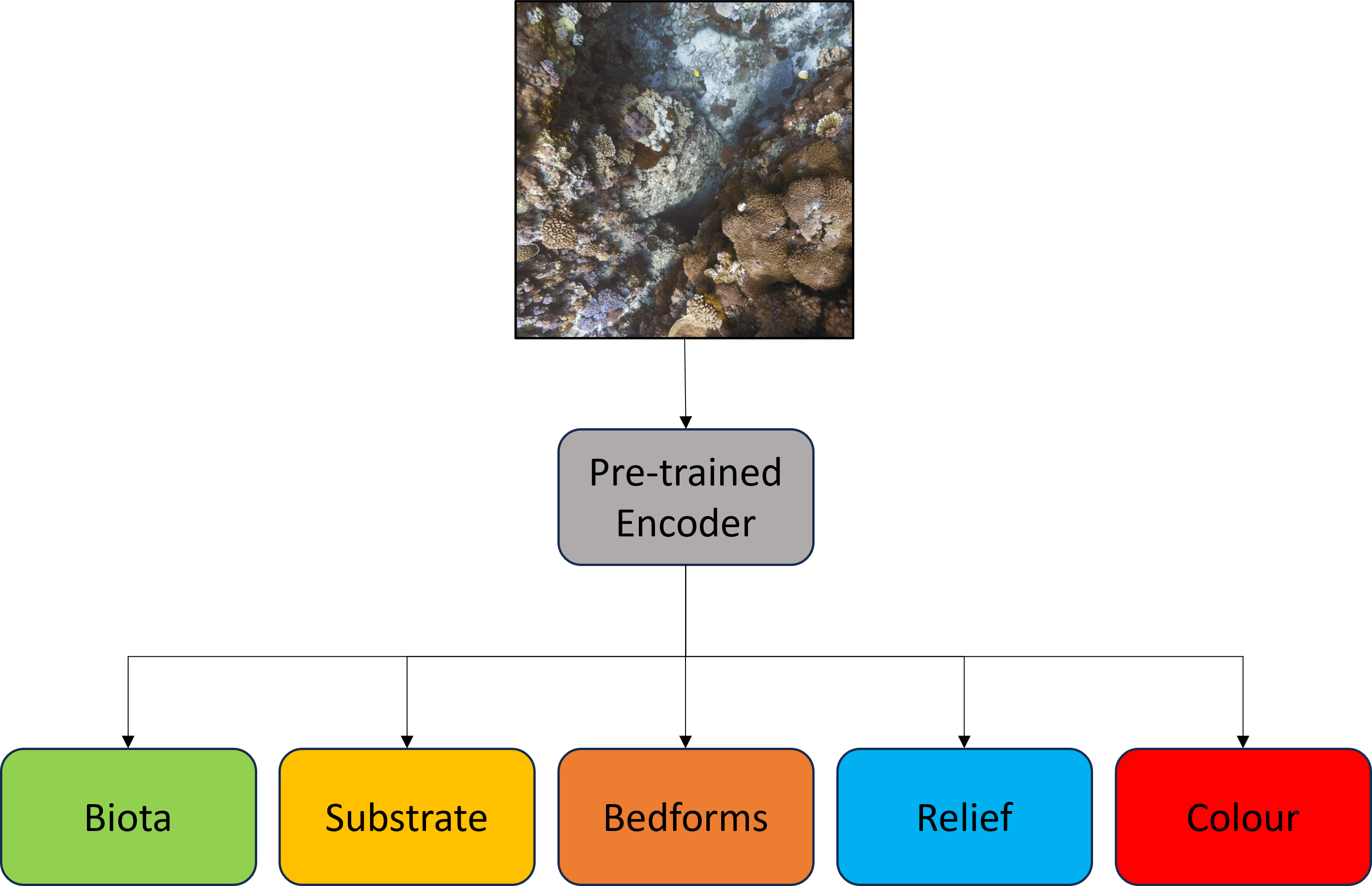}
    \caption{\textbf{Overview of CATAMI categories involved in hierarchical learning.} A pre-trained vision encoder is used to extract features from benthic imagery, which are passed to separate heads (each itself hierarchical or multi-label). The output from each head corresponds to one of the annotation categories.}
    \label{fig:catami_learning_overview}
\end{figure}

BenthicNet \cite{benthicnet} is a new dataset of seafloor photographs which aims to facilitate the development of machine learning (ML) models useful for automated benthic imagery classification.
The dataset is the largest of its kind, compiled and curated from individual datasets originating from open data repositories, government agencies, and individual research groups around the world.
The original annotations were merged by mapping them from their originating annotation schemes to a common annotation scheme (CATAMI \cite{catami}) with five independent categories (biota, substrate, bedforms, relief, and colour; see \autoref{fig:catami_learning_overview}). Of these, all but colour are hierarchical.

However, additional complexities arise due to the diversity of originating data sources.
Since each research group only annotated their data as much as was needed for their own research efforts, not only are we often missing information across different categories, but we may also be missing information necessary to exploit the full precision of the hierarchy. 
Furthermore, training a classifier requires both positive and negative examples; the ambiguity of whether an omitted label is a true negative example or simply outside the annotator's scope may complicate the use of negatively annotated samples.

\begin{figure*}[htpb]
    \centering
    \includegraphics[width=\textwidth]{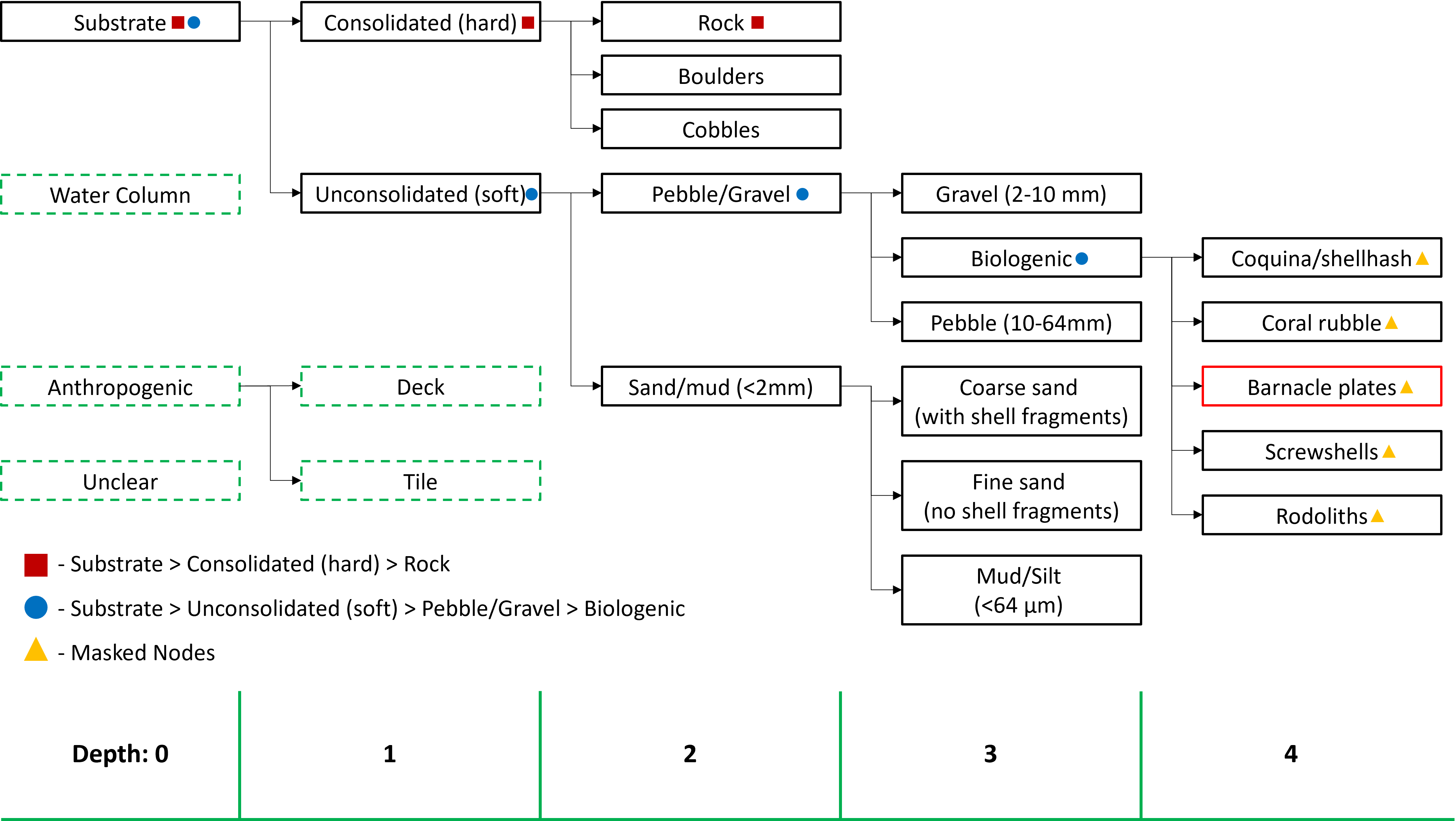}
    \caption{\textbf{CATAMI substrate label hierarchy.} The first \textit{red square} path reaches the full depth of the hierarchy along a branch extending to depth two, The second \textit{blue circle} path does not reach the leaf nodes but reaches depth three. A potential multi-label annotation is the union of these two paths, although these individual paths are also valid annotations in our scheme. The \textit{yellow triangles} illustrate the masked nodes not considered for loss, arising from a lack of precision in our \textit{blue circle} component path. Nodes with a green dashed outline are in the CATAMI-extended scheme, but not the original CATAMI scheme; a red outline denotes a node that is in the CATAMI scheme but excluded from our study due to a lack of training annotations.}
    \label{fig:catami_substrate}
\end{figure*}

The overall goal of this study is to determine how we can approach learning and evaluation in a hierarchical multi-label (HML) setting with multiple distinct forms of missing information, which can be common in real-world life sciences data. To address this goal, we establish three research questions:

\begin{enumerate}
    \item \textbf{Pre-training models.} BenthicNet is comprised of a labelled subset (12\% of images) and an unlabelled subset (88\%). Can we utilize the unlabelled data by pre-training a model using self-supervised learning (SSL)? How do state-of-the-art SSL methods compare on this task?
    \item \textbf{HML with missing information.} How can we adapt state-of-the-art (SOTA) hierarchical learning methods to an environment with multiple sources of missing information?
    \item \textbf{Model evaluation.} How can we gauge expected performance on real-world tasks, or evaluate the performance of our models in an HML setting?
\end{enumerate}

While we focus our study on BenthicNet, the real-world difficulties of incomplete annotations are common when handling amalgamated data from diverse sources.
Moreover, incompletely resolved hierarchical annotations are encountered frequently within the life sciences, since annotating plant or animal species from photographic data is challenging and the level of observable taxonomic detail can vary, even between samples within the same dataset.
Consequently, we anticipate that our methodology will be useful for a variety of fields, and its relevance is not only limited to benthic habitat imagery.

\begin{figure*}[htpb]
    \centering
    \includegraphics[width=\textwidth]{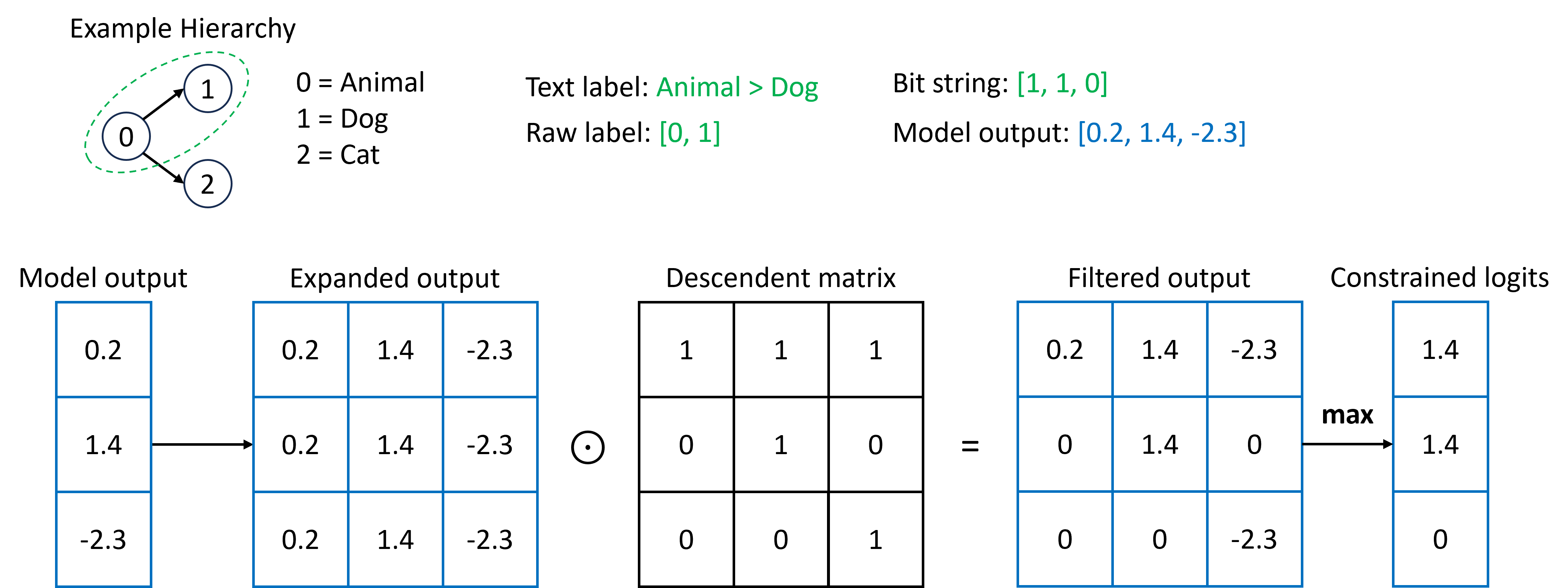}
    \caption{\textbf{Demonstration of C-HMCNN's hierarchically constrained model output \cite{coherent_hierarchical_multi-label}.} The example toy hierarchy for \texttt{animal} contains two child nodes: \texttt{dog} and \texttt{cat}. The green dotted path represents a label: \texttt{Animal $>$ Dog}. This label is also represented in its raw label form (indicating which bits are to be flipped on), and as a bit-string, where each consecutive bit represents the $n$-th node. The model output (blue) is the prediction when the image sample corresponding to the green annotation is supplied to a trained model. In this example, it violates hierarchy, since the \texttt{dog} logit is greater than the \texttt{animal} logit. This output is then expanded and filtered through the descendent (adjacency) matrix which represents our hierarchy, via the Hadamard product. Finally, the maximum along each row is taken from the filtered output to obtain a constrained model output, preserving hierarchy.}
    \label{fig:chmc_diagram}
\end{figure*}

\section{Related Work}


In ML, research on handling missing information typically focuses on methods that fill in the gaps with appropriate guesses (imputation) \cite{missing_info_survey}. Approaches such as missForest \cite{missforest}, $k$-nearest neighbours, and multiple imputations by chained equations (MICE) \cite{van_buuren_2018} generally work best with meaningful engineered features rather than raw pixels. Additionally, given the multiple types of missing information present in BenthicNet, combined with its hierarchical nature, traditional imputation approaches become highly complex. We are therefore motivated to retain our limited annotations and attempt learning despite missing information.

In the HML domain, works such as Coherent Hierarchical Multi-Label Classification Neural Networks (C-HMCNN) \cite{coherent_hierarchical_multi-label} and Hierarchical Multi-Label Classification Networks (HMCN-R) \cite{hierarchical_multi-label_class} have approached the topic via non-parameterized techniques and recurrent architectures respectively. While we leverage C-HMCNN's hierarchical constraint mechanism and loss objective, we expand beyond the scope of the original paper as we apply it to BenthicNet, with the context of multiple heads and missing information.

A number of marine studies have also studied incorporating ML into ocean-mapping. In \cite{towards_3d_benthic_maps}, Mohamed et al. explored the application of traditional machine learning methods, such as Support Vector Machines (SVM) and $k$-nearest neighbors, to imagery collected in the vicinity of Ishigaki Island (Japan), for multi-class classification. Object tracking and detection were performed on Antarctic seafloor imagery by Marini et al. \cite{automated_antarctic_benthic_fauna}, leveraging YOLO-V5 \cite{yolo-v5}. Pillay et al. applied methods such as decision trees and $k$-means to classifying bathymetry and backscatter data collected near Cape St. Francis (South Africa) \cite{benthic_mapping_ml_south_africa}. Lastly, and most similar to our work, Marine-tree \cite{marine-tree} studied hierarchical learning on a global reef dataset Reef Life Survey (RLS) --- which itself is also a component dataset in BenthicNet. In the study, Boone-Sifuentes et al. explored a variety of neural network setups and employed a modified cross-entropy loss, where the expectation across examples and hierarchical levels was used to train the model. 

To the best of our knowledge, we are the first to apply deep learning to the global scale and scope of the hierarchical and multi-label BenthicNet data and to address the task in the context of missing information. The size of this dataset additionally enables exploration of SSL, which is highly suitable in this context due to the quantity of unlabelled data. The contributions of this work can therefore be summarized as: 1) establishing methods for training hierarchical deep learning models with incomplete multi-label information; and 2) providing and benchmarking pre-trained models that are extensible and useful for downstream tasks related to underwater image classification.

\section{Background}

Our models use ResNet-50 (23.5 million parameters) \cite{resnet} and ViT-B (85.8 million parameters) \cite{vit} architectures. Both labelled and unlabelled portions of BenthicNet were used for training. The unlabelled partition was used to train a variety of encoders via SOTA SSL methods. The labelled portion was used to evaluate these encoders using probes (in which the pre-trained encoder weights are frozen and an added classifier is trained) and fine-tuning (where the pre-trained encoder weights are updated). We based our approach to HML with missing information on C-HMCNN methodology \cite{coherent_hierarchical_multi-label}.

\subsection{BenthicNet}

In this work, we use BenthicNet-1M, a curated collection of 1.35 million unannotated images, subsampled from the larger BenthicNet-11M dataset. These images depict a diverse range of global benthic habitats. Photographs were captured using a variety of methods, including by towed cameras, submersibles, and divers.
Additionally, we use the BenthicNet-Labelled dataset, which consists of 2.61 million annotations distributed across 188,688 images. This subset is split into training and test partitions based on geospatial proximity and annotation frequency. For a complete description of the construction of the BenthicNet dataset, see \cite{benthicnet}.

BenthicNet labels follow an extended CATAMI \cite{catami} scheme, developed for identifying benthic habitats. The biota category and hierarchy describes seafloor organisms, consisting of nearly 300 nodes that focus on biological morphology rather than taxonomy --- often ambiguous in underwater imagery. The substrate category describes the physical characteristics of the seafloor and contains 24 nodes. The relief category (which describes terrain height) and bedforms category (describing the shapes and patterns on the seafloor induced by physical and biological processes) consist of seven nodes each. A visualization of the substrate hierarchy is present in \autoref{fig:catami_substrate}.

With the substrate hierarchy as a visual example, we can define ``depth'' and clarify ``annotations''. We consider each level of the hierarchy as a ``depth''. In BenthicNet, an annotation can then be interpreted as a path from the root of the tree to any descendent node, with no requirement to reach a leaf node. For example, the label \texttt{Substrate $>$ Consolidated (hard)} is just as valid an annotation as \texttt{Substrate $>$ Consolidated (hard) $>$ Rock}, or even just \texttt{Substrate}. As mentioned previously, this variable precision is a type of missing information. In such cases, and as can be seen in \autoref{fig:catami_substrate}, it is possible to mask the missing descendent nodes. We elaborate on training with this kind of missing information in \autoref{sec:methodology}. 

Lastly, we note that the multi-label quality of BenthicNet implies that a valid annotation for a particular image can be any combination of the paths present in the hierarchy. To use our prior example, an image may contain pebbles and rocks, and would therefore be annotated as \texttt{Substrate $>$ Consolidated (hard) $>$ Rock, Substrate $>$ Unconsolidated (soft) $>$ Pebble/Gravel}.

\subsection{Self-supervised learning}

Given that the unlabelled data volume is more than seven times that of the labelled data, and in order to fully exploit the range of environments imaged in the dataset, SSL becomes an essential component within our training regime. In total, models were trained using six SSL methods. Each of these can be defined as either autoencoder or instance learning approaches. These SSL implementations were prepared using the \texttt{solo-learn} Python package \cite{solo-learn}.
\newline

\subsubsection{Instance and contrastive learning}

``Instance learning'' describes a family of SSL techniques that seek to train a model using multiple augmented instances of the same image. These augmented instances are referred to as ``views''. Views of each individual image are then typically passed through a student and a teacher model (self-distillation), where their outputs in embedding (or semantic) space are encouraged to be similar through a loss objective that incorporates a similarity measure \cite{SSLCookbook}.

We examine the following instance learning methods:

\begin{itemize}
    \item Simple Siamese Learning (SimSiam) \cite{simsiam};
    \item Bootstrap Your Own Latent (BYOL) \cite{byol};
    \item Barlow Twins (BT) \cite{barlow_twins}; and
    \item Momentum Contrastive Learning (MoCo-v2 and v3) \cite{mocov2, mocov3}.
\end{itemize}

Among these, SimSiam, BYOL, and BT are self-distillation methods, whereas MoCo is contrastive learning.
Contrastive learning methods are a form of metric learning, as they assert that the distance between ``positive pairs'' (formed of two views of the same data sample) should be smaller than the distance between ``negative pairs'' (formed of two views of different data samples).

Generally, these methods differ primarily in the details of post-encoder architectural implementations (employing different neural network structures such as projectors or predictors for training) and loss objectives. 
\newline

\subsubsection{Masked autoencoders}

MAE \cite{mae} is a method designed for ViTs. Much like in natural language processing (NLP), ViTs patchify an image and encode each patch of the image as a token. MAE then creates a training task similar to masked language modelling, where the model is trained to produce the full unmasked original input when provided a masked version. Most notably, MAEs have proven effective with aggressive masking (e.g., requiring 75\% of the original image to be masked).

\subsection{C-HMCNN}

Coherent hierarchical multi-label classification \cite{coherent_hierarchical_multi-label} is a SOTA hierarchical learning approach consisting of two main parts: a means to constrain the output of a model such that it does not violate hierarchical structure and a modified form of binary cross-entropy (BCE) known as max constraint loss (MCLoss), which serves as the learning objective. The constraint mechanism ensures that any output a model produces will preserve the data hierarchy (e.g., if the model predicts the presence of a Dalmatian in an image, it must therefore also be predicting the presence of a dog). The latter MCLoss ensures that the gradient direction is correct for learning. Further detail regarding this method is available in \cite{coherent_hierarchical_multi-label}.

An overview of how the constrained output is produced can be seen in \autoref{fig:chmc_diagram}. The hierarchy is first translated into a descendant matrix $\mathbf{R}$. Essentially, $\mathbf{R}$ is a graph adjacency matrix, defined such that each row and column corresponds to nodes within the hierarchy. For entry $a_{ij}$ in $\mathbf{R}$:

\[a_{ij} = \begin{cases}
    1 & \text{if node $j$ is node $i$ or a descendant of node $i$},\\
    0 & \text{otherwise.}
\end{cases}\]

Next, the model output (as logits) is expanded into an $n \times n$ matrix ($n$ being the number of hierarchy nodes), then the Hadamard product of this expanded output with $\mathbf{R}$ is calculated. This step creates a filtered output where only the descendants' logits for each row are carried over. From here, the maximum along each row is taken to obtain the constrained logits. At this point, the logits and binary label string would be evaluated with the loss criterion. 

\section{Methodology}
\label{sec:methodology}
\begin{figure*}[htpb]
    \centering
    \includegraphics[width=\textwidth]{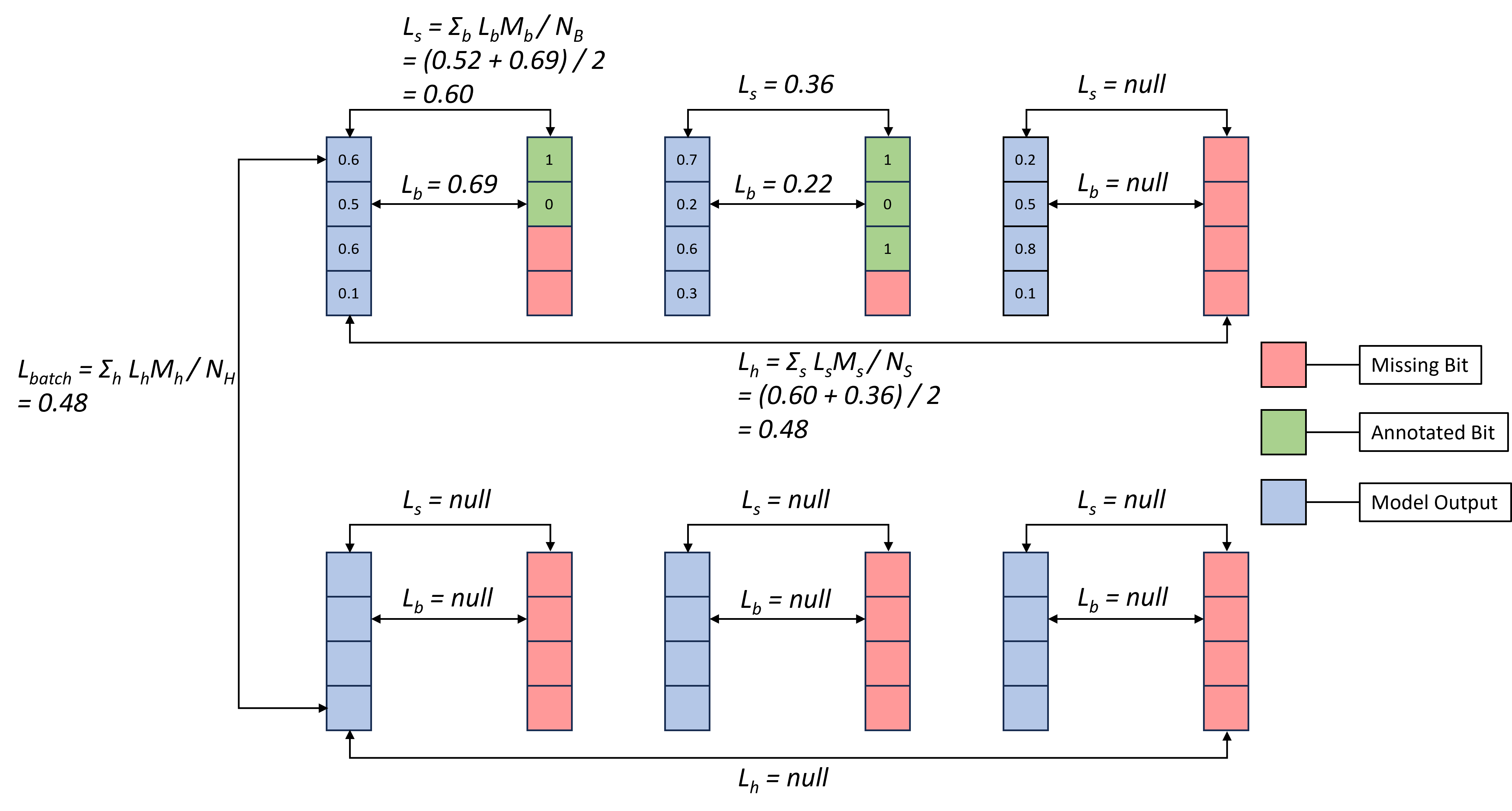}
    \caption{\textbf{Demonstration of masked loss calculation.} The example depicts how loss is calculated for two heads on a batch of size three. In this batch, the lower category is entirely missing annotations, whereas in the top category, only the last sample is missing an annotation. However, in the first two samples, for the top head, the annotations are lacking in precision. Subsequently, the loss of the entire batch only averages over the contributing green annotated bits. This philosophy is then extended to the head level.}
    \label{fig:loss_calc}
\end{figure*}

In this section, we describe our methodologies and their pertinence to our defined research questions, our approach to masking, and our evaluation methods. Our overall learning and inference pipeline consists of two main parts:

\begin{enumerate}
    \item Pre-train a model on the subsampled unlabelled portion of the dataset using SSL methods; and
    \item Extract the encoder for inference on downstream one-hot and HML tasks.
\end{enumerate}

During the downstream process, we conduct probes (linear in the case of one-hot, two-layered for HML) and fine-tuning to evaluate our encoders. The fine-tuning process is applied to the probes and their downstream classifiers in hopes of avoiding excessive destruction of learned encoder features.

\subsection{Self-supervised procedures}

\textbf{Pre-training models.} As part of answering the first research question, we initially prepared a series of self-supervised encoders from which useful encodings can be extracted. For most of the SSL methods, we use hyperparameters presented in their originating literature pertaining to ImageNet (IN-1k) \cite{imagenet}. However, in the case of Barlow Twins, we reduced the learning rate from $0.2$ by a factor of 100 to better facilitate training on BenthicNet. Unless otherwise stated, all our SSL models were trained for 100 epochs on the unlabelled BenthicNet partition.

\subsection{Hierarchical procedures}
\label{subsec:hierarchical_procedures}
Upon completing pre-training, we extracted the encoder, froze the weights (preventing further learning), and appended an architecture to enable hierarchical learning. Our hierarchical classifiers consisted of five heads, one for each CATAMI category and ``colour''. Each of our heads contained two feed-forward layers, mapping from the output dimension of the encoder to 2048, and then to the output dimension of each head (equivalent to the number of nodes present in each category). These heads were trained with 0.7 dropout. Our test set comprised around 32\% of our total labelled data and our models were trained on approximately 60\% of the data with the remainder dedicated towards the validation set.

The overall hierarchical learning process included a one-cycle cosine annealed \cite{one-cycle} learning rate with a peak of $3 \times 10^{-5}$ at the tenth epoch (start and end learning rates $3 \times 10^{-6}$). We used an AdamW optimizer \cite{adam, adamw}, 0.9 momentum, a batch size of 512, and trained for 100 epochs. Fine-tuning occurred at one-tenth of these learning rates, for 300 epochs. The fine-tuned models have therefore been trained for 400 total epochs. We used a weight decay of $1 \times 10^{-5}$ for our probes, which was reduced to zero for fine-tuning, in line with procedures presented in \cite{mae}.

\subsection{Loss calculations with missing information}
\textbf{HML with missing information.} One of the main challenges to fully utilizing the BenthicNet dataset is maximizing the use of annotated information during training despite the presence of missing information. Three different types of missing information may be distinguished:

\begin{enumerate}
    \item \textbf{Missing Precision.} Component datasets of BenthicNet do not always utilize the full depth provided in the CATAMI scheme;
    \item \textbf{Missing Branches or Sub-Trees.} Not all species or terrain characteristics are annotated; and
    \item \textbf{Missing Categories.} A sample may not have annotations for all BenthicNet categories.
\end{enumerate}

The second type of missing information, ``Missing Branches or Sub-Trees'', cannot be easily addressed without knowing the particularities of a given data source. In other words, if a node is annotated, all other unannotated nodes in that depth must be assumed to be absent in the image, even though we cannot be certain of this assumption. Otherwise, we would not have samples with negative annotations, risking a collapsed solution wherein the model only predicts positives for all nodes in the hierarchy.

The other two types of missing information, however, can be treated with masking. Our approach to masking operates similarly to how loss is computed over the number of effective tokens in NLP pre-training tasks, by taking the mean over the count of contributing non-masked components \cite{bert}.
However, the hierarchical nature of the data requires the application of masked loss across all levels of training, from the bit-wise loss between predicted and annotated nodes to the total combined losses from each categorical head. An example calculation of this hierarchical masked loss is provided in \autoref{fig:loss_calc}.

\subsection{Evaluating models}
\textbf{Model evaluation.} It can be challenging to obtain meaningful performance metrics for models trained in an HML setting. One of the first useful steps to contextualizing results is to establish a ``random'' baseline.
\newline

\subsubsection{Random baseline}
When we consider a random baseline in the case of one-hot classification, the expected performance in a balanced scenario is relatively simple: the random model predicts any of the output classes with uniform probability, resulting in a $1/n$ expected accuracy. For hierarchical scenarios, we observe that the addition of the hierarchy does not directly change this baseline. Given that we can represent the entirety of our hierarchy with $n$ nodes, the probability of the model picking any random path in the hierarchy is also $1/n$. The addition of multi-labels, however, introduces additional complexity.

In the multi-label scenario, the number of possible labels ends up becoming: \[\sum_{k=1}^{n} \binom{n}{k} = \binom{n}{1} + \binom{n}{2} + \dots + \binom{n}{n} = 2^{n}-1.\] 
Here, we increment $k$ (the number of component paths to choose) as we consider possible combinations of one, two, or more labels, up to $n$ nodes being selected. If we account for the possibility of a null label (all output bits turned off), we obtain the full set of binomial coefficients. We might at this point naïvely conclude that the expected accuracy for a random model in the HML scenario is $1/2^{n}$. However, this conclusion is premature when the hierarchical quality of the labels is considered. In HML, certain combinations of nodes in a hierarchy are not unique annotations. For example, the union of \texttt{Animal $>$ Canine $>$  Dog $>$ Dalmatian} with \texttt{Animal $>$ Canine} produces the same result as \texttt{Animal $>$ Canine $>$ Dog $>$ Dalmatian} alone. Consequently, we can imagine that $1/2^{n}$ merely acts as a lower bound for the expected accuracy score of a random model in an HML case --– only applicable when the hierarchy consists entirely of root nodes, in other words, a pure multi-label scenario.

Since the expected accuracy of a random model in an HML scenario depends on the organization of the hierarchy, a naïve approach in determining the number of valid combinatorial paths would require the evaluation of the $2^{n}$ combinations, rendering it an $\mathcal{O}(2^n)$ time-complexity operation. Subsequently, we do not explicitly calculate the expected accuracy for the HML random prediction baseline. Instead, we randomly turn on output bits and put them through the hierarchical constraint.

When we consider this Monte Carlo approach, since there are fewer nodes near the ancestral base of the hierarchy, any random combination of paths is likely to activate these lower-depth ancestor nodes. In other words, every time a node is randomly activated, that node also activates the entire path from the root of the hierarchical tree to itself due to the hierarchical constraint. Using our example animal hierarchy, even without multi-labelling, if we pick a random animal, the probability of the selection activating the canine node will be higher than the probability of the Dalmatian node being activated --- since any type of dog or wolf would also activate canine. With multi-labels, this property is exacerbated and translates to a high rate of false positives across the depths in the hierarchy. Another observation is that the largest number of possible ``states'' for combined labels occur in the middle binomial numbers ($k\!\approx\!\frac{n}{2}$). In other words, if we are randomly selecting HML annotations, the number of component paths comprising the annotation is more likely to be these middle numbers. These properties must be considered for understanding the random baseline.
\newline

\subsubsection{One-hot classification}

For model comparison and benchmarking, a simplified subset of the data was extracted for one-hot classification trials. The substrate category was selected for this purpose, and the data were filtered for images with a singular annotation at substrate depth two, which differentiates between rock, boulders, cobbles, pebble/gravel, and sand/mud. We refer to this data subset as ``Substrate-2'' \cite{benthicnet}.
In addition to Substrate-2, we also investigated one of the source datasets, ``German Bank 2010'', using the original labels: silt/mud, silt with bedforms, reef, glacial till, and sand with bedforms \cite{german2010}. German Bank 2010 is a valuable and realistic example of a typical benthic image dataset that may be encountered in practice. In total, Substrate-2 contains 75,537 images, whilst German Bank 2010 is notably smaller at 3,181 images.

In the one-hot context, we use macro-averaged evaluations (instead of sample-weighted), which account for model performance across imbalanced classes. Here we record the macro-average $F_1$ score. For all one-hot probes, we used a linear classifier and a batch size of 1024. The remaining hyperparameters were as described in our hierarchical procedures (\autoref{subsec:hierarchical_procedures}). However, for fine-tuning ViTs, we found that a one-cycle cosine-annealing learning rate scheduler with a peak of $3 \times 10^{-7}$ and an initial/final learning rate of $3 \times 10^{-8}$ produced more stable results.
\newline

\subsection{Evaluating HML without imbalance}

HML models are typically evaluated using metrics that summarize the trade-off between precision and recall, such as the area under the precision-recall curve (AUPRC) \cite{coherent_hierarchical_multi-label, hierarchical_multi-label_class} or the average precision (AP), defined as:

\[ \operatorname{AP} = \sum_{n} (R_{n}-R_{n-1}) \, P_{n}.\]

This definition for AP is the rectangular approximation of AUPRC \cite{scikit-learn}. In this implementation, the thresholds that determine $n$ are set by the number of unique values in the prediction array, with each successive threshold enabling more components of the prediction array to be classified as ``positive'', creating the approximated precision-recall curve.

While the AP summation should consist of a variable $n$ terms, in practice $n=2$, as only a single threshold at 0.5 is used to binarize the output probabilities into presence/absence predictions bits for sigmoid outputs, which is then fed into the AP calculation. The reason we use this binarized output rather than the sigmoid output for evaluation is due to the sometimes overly optimistic scores generated for the latter.\footnote{A detailed discussion of this topic is provided with our code repository.}

When we take the AP score, we set ``micro'' as the averaging method. Intuitively, ``micro-averaging'' provides an idea of the ``correctness'' for each predicted bit, while the default ``macro-average'' relays the ``correctness'' for each sample. When this metric is applied to the entire test set, we refer to it as just the ``AP'' score. At this point, imbalance across annotations or depths is not considered.
\newline

\subsubsection{Evaluating HML with imbalance}

In addition to an overall AP score for the model, it is informative to explore how the models performed when considering imbalance across labels and depths of the hierarchies. Imbalance is considered in two ways:

\begin{enumerate}
    \item Each unique HML annotation is considered and evaluated independently using the AP score, and results are macro-averaged across the annotations within a depth, then across the depths (HML AP); and
    \item Each annotation is decomposed into its component paths, and then macro-averaged across these component annotations within each depth, then across the depths (Singular $F_1$).
\end{enumerate}
We found that each evaluation method provided a complementary understanding of what the models have learnt.

\section{Results}

\begin{table*}[ht]
\centering
\caption{Accuracy and macro-averaged $F_1$ score (\%) for linear probing and fine-tuning pre-trained encoders on one-hot datasets Substrate-2 and German Bank 2010. \textbf{Bold:} best ResNet-50 average score for a given evaluation. For comparison, we also show the performance of a model trained end-to-end from scratch for the same number of steps (None (Rand. init.)).}
\label{tab:one_hot_metrics}
\begin{tabular}{llcccccccc}
\toprule
& & \multicolumn{4}{c}{Substrate-2} & \multicolumn{4}{c}{German Bank 2010} \\
\cmidrule(lr){3-6} \cmidrule(lr){7-10}
& & \multicolumn{2}{c}{Linear probe} & \multicolumn{2}{c}{Fine-tune} & \multicolumn{2}{c}{Linear probe} & \multicolumn{2}{c}{Fine-tune} \\
\cmidrule(lr){3-4} \cmidrule(lr){5-6} \cmidrule(lr){7-8} \cmidrule(lr){9-10}
Backbone & Pre-training & Accuracy & $F_1$ & Accuracy & $F_1$ & Accuracy & $F_1$ & Accuracy & $F_1$ \\
\midrule
ResNet-50
& None (Rand. init.)
    & 76.4\terr{0.4} 
    & \textbf{53.3}\terr{0.7} 
    & 78.7\terr{0.3} 
    & 57.9\terr{0.8} 
    & 53.4\terr{2.4} 
    & 43.0\terr{3.8} 
    & 54.1\terr{1.0} 
    & 46.7\terr{3.2} \\
& IN-1k Supervised 
    & 75.5\terr{0.3} 
    & 51.2\terr{0.5} 
    & \textbf{84.0}\terr{0.2} 
    & \textbf{66.6}\terr{0.4} 
    & 37.6\terr{5.2} 
    & 30.0\terr{2.9} 
    & 65.9\terr{4.0} 
    & 59.2\terr{4.2} \\
& BT-400ep 
    & \textbf{76.8}\terr{0.2} 
    & 51.4\terr{0.4} 
    & 82.8\terr{0.1} 
    & 64.0\terr{0.1} 
    & \textbf{55.9}\terr{2.4} 
    & \textbf{43.2}\terr{6.0} 
    & \textbf{77.0}\terr{0.7} 
    & \textbf{72.3}\terr{0.8} \\
\midrule
ViT-B
& MoCo-v3 
    & 81.2\terr{0.0} 
    & 58.8\terr{0.5} 
    & 85.5\terr{0.3} 
    & 68.7\terr{0.8} 
    & 57.9\terr{6.5} 
    & 49.1\terr{10.7} 
    & 77.4\terr{0.2} 
    & 70.0\terr{2.5} \\
\bottomrule
\end{tabular}
\end{table*}

\begin{table*}[tb]
\centering
\caption{Evaluations of pre-trained models on substrate classification in HML setting. \textbf{Bold:} best ResNet-50 and ViT-B average scores for a given evaluation. For comparison, we also show the performance that would be obtained with random prediction, and of a ResNet-50 model trained end-to-end from scratch for the same number of steps (None (Rand. init.)).}
\label{tab:results_overall_scores}
\begin{tabular}{llcccccc}
\toprule
& & \multicolumn{2}{c}{Average precision (AP)} & \multicolumn{2}{c}{HML AP} & \multicolumn{2}{c}{Singular $F_1$} \\
\cmidrule(lr){3-4} \cmidrule(lr){5-6} \cmidrule(lr){7-8}
Backbone & Pre-training & Probe & Fine-tune & Probe & Fine-tune & Probe & Fine-tune \\
\midrule
\textit{None}
& Random Prediction
    & 21.22\terr{0.02} 
    & 21.22\terr{0.02} 
    & 43.78\terr{0.74} 
    & 43.78\terr{0.74} 
    & 28.88\terr{0.12} 
    & 28.88\terr{0.12} \\
\midrule
ResNet-50
& None (Rand. init.)
    & 66.37\terr{0.61} 
    & 69.19\terr{0.46} 
    & 63.06\terr{1.37} 
    & 64.45\terr{0.69} 
    & 18.92\terr{1.00} 
    & 24.35\terr{1.41} \\
& IN-1k Supervised
    & 72.16\terr{0.22} 
    & 71.29\terr{0.12}
    & 64.45\terr{0.30} 
    & 62.78\terr{0.58}
    & 26.26\terr{1.17} 
    & 24.24\terr{1.16} \\
& BT-100ep 
    & \textbf{73.32}\terr{0.25} 
    & \textbf{71.93}\terr{0.32}
    & 67.29\terr{1.39} 
    & \textbf{65.67}\terr{1.41} 
    & \textbf{33.61}\terr{5.33} 
    & 32.31\terr{3.75} \\
& BYOL 
    & 73.01\terr{0.22} 
    & 71.61\terr{0.08}
    & \textbf{67.52}\terr{2.23} 
    & 65.39\terr{1.87} 
    & 32.89\terr{4.74} 
    & \textbf{34.41}\terr{2.52} \\
& MoCo-v2 
    & 69.93\terr{0.13} 
    & 70.89\terr{0.28} 
    & 63.36\terr{0.10} 
    & 65.33\terr{1.65} 
    & 22.90\terr{1.41} 
    & 28.90\terr{3.96} \\
& SimSiam 
    & 72.18\terr{0.18} 
    & 68.70\terr{0.08} 
    & 67.03\terr{1.86} 
    & 61.16\terr{0.13} 
    & 29.17\terr{4.85} 
    & 23.58\terr{0.96} \\
\midrule
ViT-B
& MAE 
    & 74.35\terr{0.47} 
    & 73.51\terr{0.41} 
    & 63.72\terr{0.63} 
    & 63.43\terr{0.78} 
    & 25.18\terr{1.59} 
    & 26.12\terr{0.13} \\
& MoCo-v3 
    & \textbf{77.36}\terr{0.46} 
    & \textbf{76.36}\terr{0.31} 
    & \textbf{65.06}\terr{0.13} 
    & \textbf{64.31}\terr{0.40} 
    & \textbf{32.51}\terr{2.16} 
    & \textbf{33.24}\terr{2.65}
 \\
\bottomrule
\end{tabular}
\end{table*}

\begin{figure*}[htbp]
    \centering
    \includegraphics[width=\textwidth]{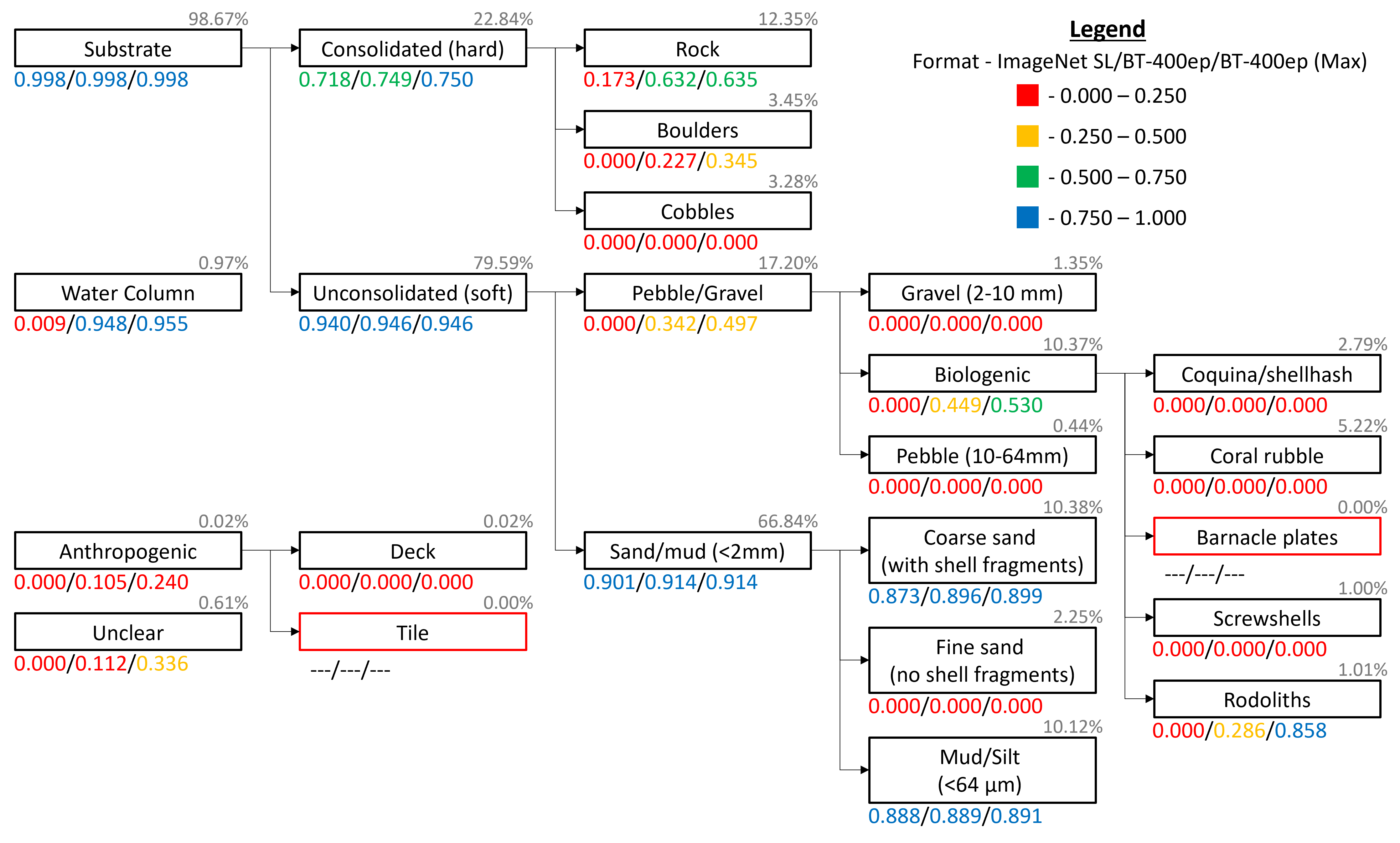}
    \caption{\textbf{Model performance per substrate node.} Detailed below each node are the average $F_1$ scores in the singular node case, comparing a supervised ImageNet pre-trained encoder to a BT encoder pre-trained for 400 epochs on BenthicNet (BT-400ep). The third score, BT-400ep (Max), represents the maximum the BT-400ep was able to achieve across three trials. The light grey percentage in the top right corner shows the proportion of samples that contain a positive instance of the node over all hierarchical substrate test data. Lastly, while the training set contains samples of \texttt{Anthropogenic $>$ Tile}, the test set does not. Neither training nor test sets contain barnacle plate samples.}
    \label{fig:node-wise_f1s}
\end{figure*}

In \autoref{tab:one_hot_metrics}, we present the results for one-hot classification from \cite{benthicnet}, using the Substrate-2 and German Bank 2010 \cite{german2010} datasets, compared against novel results for our strongest performing ViTs (MoCo-v3). Here, we compare the randomly initialized baseline (Rand. Init.), against a PyTorch encoder pre-trained in a supervised manner on ImageNet (IN-1k) \cite{imagenet} for 600 epochs \cite{pytorch}, and a BT model pre-trained for 400 epochs (BT-400ep) on the unlabelled BenthicNet dataset.

The average hierarchical results for the complete dataset, HML, and singular cases are shown in \autoref{tab:results_overall_scores}. In addition to the pre-trained SSL models, we also include results for random predictions (Rand. Pred.), randomly initialized (Rand init.) models trained for 100 and 400 epochs respectively with unfrozen encoder weights (to match probes and fine-tuning models), and ImageNet supervised. In this case, the BT model was trained for 100 epochs (BT-100ep) in line with other SSL models, and explicitly distinguished from its 400 epoch counterpart used in our other presented results. As in the one-hot case, our probe and fine-tuned results include the average and standard deviation across three trials with random seeds.

In \autoref{fig:node-wise_f1s}, we present a deeper hierarchical view of the singular $F_1$ results in \autoref{fig:node-wise_f1s}, indicating the average and maximum results for BT-400ep compared to ImageNet Sup. 

\section{Discussion}

While our SSL-based encoders offer competitive results to the IN-1k supervised baseline despite fewer pre-training epochs, our results suggest only minor effects resulting from the pre-training method. The primary utility of these results is to provide benchmarks for the self-supervised models we intend to share publicly. We also note that some fine-tuned models indicate no significant improvement over their respective probes (at least not consistently across all metrics), which could potentially be addressed by tailoring the fine-tuning procedure according to each individual model.

In the one-hot scenario, while standard approaches that apply an ImageNet pre-trained encoder may provide competitive results for datasets with abundant samples such as our Substrate-2 subset, they perform notably worse than the pre-trained SSL model for the much smaller German Bank 2010 subset. The size of German Bank 2010 is typical of many local or regional benthic environmental image datasets, which we believe offers a practical case study for these models. The large discrepancy in performance perhaps results from the pre-training domains for these models --- though ImageNet does contain some underwater imagery and benthic biota classes, it is not its focus \cite{imagenet}.

Although our metrics evaluate the same predictions, some apparently contradictory results were observed. We note that the AP scores, which measure similarity between the output and the labelled bit-strings, tend to show high performance for trained models and low performance for random predictions. On the other hand, when we look at our decomposed metric, we find that the random baseline is now much higher relative to trained models. As noted earlier, the random baseline tends to select multi-label annotations. Hence, this property leads to a high degree of false positives. This property results in lower scores for the random bit-wise baseline. Since trained models do not make as many false positive predictions, they are not as heavily penalized in this context. 
Generally speaking, in a deep and sparse hierarchy containing many nodes, the bit-strings for our labels become sparse. Subsequently, the difference between a model making many false positive guesses and a model trained on the HML task also becomes more distinct. Additionally, partially correct answers are scored higher, since more of the zeros in the bit-string will be matching up. This sparsity effect motivates examining model performance on each node individually, as we are also interested in the nuances of our model predictions.

\autoref{fig:node-wise_f1s} provides insight into what the models are learning. Since we are evaluating predictions for each node, we move away from a bit-wise metric and resort to a standard $F_1$ score, defined by precision and recall. We note that despite the overall $F_1$ scores, the precision and recall patterns for the random baseline and trained models are quite distinct. The random baseline typically has low precision but high recall, while trained models have a more balanced precision and recall. Again, the cause of this behaviour is due to the random baseline picking up a high number of false positives. 

As we can see from \autoref{fig:node-wise_f1s}, trained models do reasonably well at earlier depths, before falling off dramatically in later ones. However, the $F_1$ scores at later depths tend to be zero. These scores are not possible if the models were confusing deep nodes for one another. Instead, we confirm in our training logs that learned models do not seem to be making predictions to this level at all. As an example, using our substrate hierarchy depicted in \autoref{fig:node-wise_f1s}, when provided an image containing coral rubble, the model will predict the image to a lower-depth level at biologenic, or pebble/gravel, rather than to the full depth of the hierarchy. In other words, trained models become less confident at deeper hierarchical levels. These ``partial predictions'' are the same that receive high scores in the multi-label bit-wise cases. 

A notable exception to this behaviour is in the case of ``rhodoliths''. In one of our three trials, the BT-400ep probe appeared to have learned the signal for rhodoliths as interpreted by the encoder. The $F_1$ score was relatively strong in this case. This phenomenon seems to suggest that although the encoder does indeed detect signals for nodes at a high hierarchical depth, the classifier is not consistently picking up on these signals. Hence, it is also informative to view the maximum scores across all trials. Additionally, we note that rhodoliths typically have a distinct reddish hue, which may make them easier to recognize. This observation also potentially suggests that the encoders may be able to distinguish between deeper level nodes, but due to the imbalance between positive and negative instances for these nodes seen during training, they make predictions that fall below the threshold of 0.5, failing to qualify it as a true positive prediction.

\section{Conclusion and Future Work}

In this work, we demonstrate the application of SOTA self-supervised pre-training methods for benthic imagery transfer learning and classification. We also present a means to adapt the HML learning framework for a scenario comprising real-world data with multiple types of missing information.

We found that while standard supervised ImageNet encoders performed strongly on datasets with abundant samples, they may experience poorer results compared to self-supervised BenthicNet encoders when applied to small datasets common in ocean sciences. Additionally, we found that in the HML setting, self-supervised encoders appeared to extend deeper into the hierarchy when classifying results.

One complicating factor is hierarchical imbalance. Due to the nature of real-world HML datasets such as BenthicNet, there may exist a notable imbalance between the nodes within the hierarchy, which may also dependent upon the depth of the node. Accounting for imbalance during training is non-trivial as loss weighting or weighted sampling may not be feasible as a result of the multi-labelled data. Instead, a node or bit-level weighting for positive instances of the node may help produce stronger hierarchical probes or fine-tuned models.

Another immediately feasible task is in the examination of the other CATAMI categories. While biota is no doubt of interest to many ocean scientists, cataloguing and visualizing the details of model performance on the hierarchy is challenging due to its sheer size. However, we believe we can reasonably approach this category by breaking the biota hierarchy into sub-trees which may then be studied individually.

\section{Acknowledgements}
This research was supported with funding provided by the Ocean Frontier Institute (\href{https://www.ofi.ca/}{OFI}) as part of the Benthic Ecosystem Mapping and Engagement (\href{https://www.ofibecome.org/}{BEcoME}) project, through an award from the Canada First Research Excellence Fund. Compute resources support was provided by \href{https://ace-net.ca/}{ACENET} and the Digital Research Alliance of Canada (\href{https://alliancecan.ca/}{DRAC}).

\FloatBarrier

\bibliographystyle{IEEEtran}
\bibliography{references}

\FloatBarrier

\end{document}